\documentclass{article}

\PassOptionsToPackage{numbers, compress}{natbib}

     \usepackage[final]{neurips_2019}

\usepackage[utf8]{inputenc} 
\usepackage[T1]{fontenc}    
\usepackage{hyperref}       
\usepackage{url}            
\usepackage{booktabs}       
\usepackage{amsfonts}       
\usepackage{nicefrac}       
\usepackage{microtype}      
\usepackage{amsmath}
\usepackage{amsfonts}
\usepackage{amssymb}
\usepackage{amsthm}
\usepackage{algorithm}
\usepackage[noend]{algpseudocode}
\usepackage{graphicx}
\usepackage{subcaption}
\usepackage{natbib}
\PassOptionsToPackage{options}{natbib}

\title{Deep Bayesian Reward Learning from Preferences}

\author{%
  Daniel S. Brown\\
  Department of Computer Science\\
  University of Texas at Austin\\
  \texttt{dsbrown@cs.utexas.edu} \\
   \And
  Scott Niekum\\
  Department of Computer Science\\
  University of Texas at Austin\\
  \texttt{sniekum@cs.utexas.edu} \\
}

\begin{document}

\maketitle

\begin{abstract}
Bayesian inverse reinforcement learning (IRL) methods are ideal for safe imitation learning, as they allow a learning agent to reason about reward uncertainty and the safety of a learned policy. However, Bayesian IRL is computationally intractable for high-dimensional problems because each sample from the posterior requires solving an entire Markov Decision Process (MDP). While there exist non-Bayesian deep IRL methods, these methods typically infer point estimates of reward functions, precluding rigorous safety and uncertainty analysis. We propose Bayesian Reward Extrapolation (B-REX), a highly efficient, preference-based Bayesian reward learning algorithm that scales to high-dimensional, visual control tasks. Our approach uses successor feature representations and preferences over demonstrations to efficiently generate samples from the posterior distribution over the demonstrator's reward function without requiring an MDP solver. Using samples from the posterior, we demonstrate how to calculate high-confidence bounds on policy performance in the imitation learning setting, in which the ground-truth reward function is unknown. We evaluate our proposed approach on the task of learning to play Atari games via imitation learning from pixel inputs, with no access to the game score. We demonstrate that B-REX learns imitation policies that are competitive with a state-of-the-art deep imitation learning method that only learns a point estimate of the reward function. Furthermore, we demonstrate that samples from the posterior generated via B-REX can be used to compute high-confidence performance bounds for a variety of evaluation policies. We show that high-confidence performance bounds are useful for accurately ranking different evaluation policies when the reward function is unknown. We also demonstrate that high-confidence performance bounds may be useful for detecting reward hacking.
\end{abstract}

\section{Introduction}
As robots and other autonomous agents enter our homes, schools, workplaces, and hospitals, it is important that these agents can safely learn from and adapt to a variety of human preferences and goals. One common way to learn preferences and goals is via imitation learning, in which an autonomous agent learns how to perform a task by observing demonstrations of the task. While there exists a large body of literature on high-confidence off-policy evaluation in the reinforcement learning (RL) setting, there has been much less work on high-confidence policy evaluation in the imitation learning setting where reward samples are unavailable. 

Prior work on high-confidence policy evaluation for imitation learning has used Bayesian inverse reinforcement learning (IRL) \cite{ramachandran2007bayesian} to allow an agent to reason about reward uncertainty and policy robustness \cite{brown2018efficient,brown2018risk}. However, Bayesian IRL is typically intractable for complex problems due to the need to repeatedly solve an MDP in the inner loop. This high computational cost precludes robust safety and uncertainty analysis for imitation learning in complex high-dimensional problems.

We first formalize the problem of high-confidence off-policy evaluation \cite{thomas2015high} for imitation learning \cite{Argall2009}. We next propose a novel algorithm, Bayesian Reward Extrapolation (B-REX), 
that uses a pairwise ranking likelihood to significantly reduce the computational complexity of generating samples from the posterior distribution over reward functions when performing Bayesian IRL. We demonstrate that B-REX can leverage neural network function approximation and successor features \cite{barreto2017successor} to efficiently perform deep Bayesian reward inference given preferences over demonstrations that consist of raw visual observations. Finally, we demonstrate that samples obtained from B-REX can be used to solve the high-confidence off-policy evaluation problem for imitation learning in high-dimensional tasks. We evaluate our method on imitation learning for Atari games and demonstrate that we can efficiently compute high-confidence bounds on the worst-case performance of a policy and that these bounds are beneficial when comparing different evaluation policies and may provide a useful tool for detecting reward hacking \cite{amodei2016concrete}.

\section{Related work}

Imitation learning is the problem of learning a policy from demonstrations of desired behavior. Imitation learning can roughly be divided into techniques that use behavioral cloning and techniques that use inverse reinforcement learning. Behavioral cloning methods \cite{pomerleau1991efficient} seek to solve the imitation learning problem via supervised learning where the goal is to learn a mapping from states to actions that mimics the demonstrator. While computationally efficient, these methods can suffer from compounding errors \cite{ross2011reduction}. Methods such as DAgger \cite{ross2011reduction} and DART \cite{laskey2017dart} avoid this problem by collecting additional state-action pairs from a demonstrator in an online fashion.

Inverse reinforcement learning methods \cite{ng2000algorithms} typically seek to solve the imitation learning problem by first estimating a reward function that makes the demonstrations appear near optimal and then performing reinforcement learning \cite{sutton1998introduction} on the inferred reward function to learn a policy that can generalize to states not seen in the demonstrations. Classical approaches typically repeatedly alternate between reward estimation and full policy optimization. Bayesian IRL \cite{ramachandran2007bayesian} generates samples from the posterior distribution over rewards, whereas other methods seek a single estimate of the reward that matches the demonstrator's state occupancy \cite{abbeel2004apprenticeship}, often while also seeking to maximize the entropy of the resulting policy \cite{ziebart2008maximum}. 
Modern, deep learning approaches to inverse reinforcement learning are typically based on a maximum entropy framework \cite{finn2016guided} or an occupancy matching framework \cite{ho2016generative} and are related to Generative Adversarial Networks \cite{goodfellow2014generative,finn2016connection}. These methods scale to complex control problems by iterating between reward learning and policy learning steps. Recently, Brown et al. proposed to use  preferences over suboptimal demonstrations to efficiently learn a reward function via supervised learning without requiring fully or partially solving an MDP \cite{browngoo2019trex,brown2019drex}. The reward function is then used to optimize a potentially better-than-demonstrator policy. However, despite recent successes of deep IRL, existing methods typically return a point estimate of the reward function, precluding the rich uncertainty and robustness analysis possible with a full Bayesian approach. One of our contributions is to propose the algorithm B-REX, first deep Bayesian IRL algorithm that can scale to complex control problems with visual observations.

Another contribution of this paper is an application of B-REX to safe imitation learning \cite{brown2018efficient}.
While there has been much recent interest and progress in imitation learning \cite{arora2018survey}, less attention has been given to problems related to safe imitation learning. Zhang and Cho propose SafeDAgger \cite{safedagger} a variant of DAgger that predicts in which states the novice policy will have a large action difference from the expert policy. Control is given the the expert policy only if the predicted action difference of the novice is above some hand-tuned parameter, $\tau$. Other work has focused making generative adversarial imitation learning \cite{ho2016generative} more robust and risk-sensitive.  Lacotte et al. \cite{lacotte2019risk} propose an imitation learning algorithm that seeks to match the tail risk of the expert as well as find a policy that is indistinguishable from the demonstrations. Brown and Niekum \cite{brown2018efficient} propose a Bayesian sampling approach to provide explicit high-confidence safety bounds in the imitation learning setting. Their method uses samples from the posterior distribution $P(R|D)$ to compute sample efficient probabilistic upper bounds on the policy loss of any evaluation policy. Brown et al. \cite{brown2018risk} extend this work by proposing an active learning algorithm that uses these high-confidence performance bounds for risk-aware policy improvement via active queries. Our work presented in this paper extends and generalizes the work of Brown and Niekum \cite{brown2018efficient} by demonstrating, for the first time, that high-confidence performance bounds can be obtained for imitation learning problems where demonstrations consist of high-dimensional visual observations.


Safety has been extensively studied within the reinforcement learning community (see Garcia et al. \cite{garcia2015comprehensive} for a survey). These approaches usually either seek safe exploration strategies or seek to optimize an objective other than expected return. Recently, objectives based on measures of risk such as VaR and Conditional VaR have been shown to provide tractable and useful risk-sensitive measures of performance for MDPs \cite{tamar2015optimizing,chow2015risk}. Other related work on safe reinforcement learning has focused finding robust solutions to MDPs using Bayesian ambiguity sets \cite{petrik2019beyond} and on obtaining high-confidence off-policy bounds on the performance of an evaluation policy \cite{thomas2015high,hanna2017bootstrapping}. Recently, it has been shown that high-confidence off policy evaluation is possible when samples of the true reward are available but the behavior policy is unknown \cite{hanna2019importance}. Our work complements existing work on high-confidence off policy evaluation by formulating and providing a deep learning solution to the problem of high-confidence off-policy evaluation in the imitation learning setting, where samples of rewards are \textit{not observed} and the demonstrator's policy (the behavioral policy) is unknown. 

\section{Preliminaries}

\subsection{Notation}
We model the environment as a Markov Decision Process (MDP) consisting of states $\mathcal{S}$, actions $\mathcal{A}$, transition dynamics $T:S \times A \times S \to [0,1]$, reward function $R:S \to \mathbb{R}$, initial state distribution $S_0$, and discount factor $\gamma$. A policy $\pi$ is a mapping from states to a probability distribution over actions. We denote the value of a policy $\pi$ under reward function $R$ as $V^\pi_R = \mathbb{E}_{s_0 \sim S_0}[\sum_{t=0}^\infty \gamma^t R(s_t) | \pi]$ and denote the value of executing policy $\pi$ starting at state $s \in S$ as 
$V^\pi_R(s) = \mathbb{E}[\sum_{t=0}^\infty \gamma^t R(s_t) | \pi, s_0 = s]$. 
Given a reward function $R$, we denote the Q-value of a state-action pair $(s,a)$ as $Q^{\pi}_R(s,a) = R(s) + \gamma \sum_{s' \in S} T(s,a,s') V^\pi_R(s')$. We use the notation $V^*_R = \max_{\pi} V^\pi_R$ and $Q^*_R(s,a) = \max_{\pi} Q^{\pi}_R(s,a)$.

\subsection{Bayesian Inverse Reinforcement Learning}
In inverse reinforcement learning, the environment is modeled as an MDP$\setminus$R where the reward function $R$ is internal to the demonstrator and is unknown and unobserved by the learner. The goal of inverse reinforcement learning (IRL) is to infer the latent reward function of the demonstrator given demonstrations consisting of state-action pairs from the demonstrator's policy. Bayesian IRL models the demonstrator as a Boltzman rational agent that  follows the softmax policy
\begin{equation}
\pi_R(a|s) = \frac{e^{\beta Q^*_R(s,a)}}{\sum_{b \in \mathcal{A}} e^{\beta Q^*_R(s,b)}},
\end{equation}
where $R$ is the reward function of the demonstrator, and $\beta \in [0, \infty)$ is the inverse temperature parameter that represents the confidence that the demonstrator is acting optimally.
Given the assumption of Boltzman rationality, the likelihood of a set of demonstrations 
$D = \{ (s,a) : (s,a) \sim \pi_D \}$,
given a specific reward function hypothesis $R$, can be written as 
\begin{equation} \label{eqn:boltzman}
P(D | R) = \prod_{(s,a) \in D} \pi_R(a|s) = \prod_{(s,a) \in D} \frac{e^{\beta Q^*_R(s,a)}}{\sum_{b \in \mathcal{A}} e^{\beta Q^*_R(s,b)}}.
\end{equation}

Bayesian IRL \cite{ramachandran2007bayesian}  generates samples from the posterior distribution
$P(R|D) \sim P(D|R)P(R)$
via Markov Chain Monte Carlo (MCMC) sampling. This requires repeatedly solving for $Q^*_R$ in order to compute the likelihood of each new proposal. Thus, Bayesian IRL methods are typically only used for low-dimensional problems with reward functions that are often linear combinations of a small number of hand-crafted features \cite{brown2018efficient,biyik2019asking}. One of our contributions is to propose an efficient deep Bayesian reward learning algorithm that leverages preferences to allow Bayesian IRL to be scaled to high-dimensional visual control problems.

\section{High Confidence Off-Policy Evaluation for Imitation Learning} \label{sec:hcope-il}
Before detailing B-REX, we first formalize the problem of high-confidence off-policy evaluation for imitation learning. We assume an MDP$\setminus$R, an evaluation policy $\pi_{\rm eval}$, a set of demonstrations, $D = \{(s_1,a_1),\ldots,(s_m,a_m)\}$, confidence level $\delta$, and performance statistic $g:\Pi \times \mathcal{R} \rightarrow \mathbb{R}$, where $\mathcal{R}$ denotes the space of all reward functions and $\Pi$ is the space of all policies.

The \textit{High-Confidence Off-Policy Evaluation problem for Imitation Learning} (HCOPE-IL) is to find a high-confidence lower bound $\hat{g}: \Pi \times \mathcal{D}$ such that $\text{Pr}(g(\pi_{\rm eval}, R^*) \geq \hat{g}(\pi_{\rm eval}, D)) \geq 1 - \delta$, where $R^*$ denotes the demonstrator's true reward function, and $\mathcal{D}$ denotes the space of all demonstration sets $D$. HCOPE-IL takes as input an evaluation policy $\pi_{\rm eval}$, a set of demonstrations $D$, and a performance statistic, $g(\pi)$, which evaluates a policy under a reward function. The goal of HCOPE-IL is to return a high-confidence lower bound $\hat{g}$ on the performance statistic $g(\pi_{\rm eval}, R^*)$.

Note that this problem setting is significantly more challenging than the standard high-confidence off-policy evaluation problem in reinforcement learning, which we denote as HCOPE-RL. In HCOPE-RL the behavior policy is typically known and the demonstrations from the behavior policy contain ground-truth reward samples \cite{thomas2015high}. In HCOPE-IL, the behavior policy is the demonstrator's policy $\pi_{R^*}$, which is unknown. Furthermore, in HCOPE-IL the demonstration data from $\pi_{R^*}$ contains only state-action pairs; samples of the true reward signal are not available. In the following sections we describe how to use preferences to scale Bayesian IRL to high-dimensional visual control tasks as a way to efficiently solve the HCOPE-IL problem for complex, visual imitation learning tasks.

\section{Deep Bayesian Reward Extrapolation}
Prior work \cite{brown2018efficient,brown2018risk} has investigated HCOPE-IL for simple problem domains where repeatedly solving for optimal Q-values is possible. However, for high-dimensional tasks such as learning control policies from pixel observations, even solving a single MDP can be challenging and sampling from $P(R|D)$ becomes intractable. We now describe one of the main contribution of this paper: scaling Bayesian IRL to high-dimensional visual imitation learning problems.

Our first insight towards solving this problem is that the main bottleneck for standard Bayesian IRL \cite{ramachandran2007bayesian} is computing the softmax likelihood function:
\begin{equation}
P(D | R) = \prod_{(s,a) \in D} \frac{e^{\beta Q^*_R(s,a)}}{\sum_{b \in \mathcal{A}} e^{\beta Q^*_R(s,b)}}.
\end{equation}
which requires solving for optimal Q-values. Thus, to make Bayesian IRL scale to high-dimensional visual domains, it is necessary to either efficiently solve for optimal Q-values or to formulate a new likelihood. Value-based reinforcement learning focuses on solving for optimal Q-values quickly; however, even for low-resolution visual control tasks such as Atari, RL algorithms can several hours or even days to train \cite{mnih2015human,hessel2018rainbow}. Because MCMC is sequential in nature, evaluating large numbers of proposal steps is infeasible given the current state-of-the-art in RL. Methods such as transfer learning could reduce the time needed to calculate $Q^*_R$ for a new proposed reward $R$; however, transfer learning is not guaranteed to speed up reinforcement learning on the new task \cite{taylor2009transfer} and transfer learning methods that avoid performing reinforcement learning only provide loose bounds on policy performance \cite{barreto2017successor}, making it difficult to compute accurate likelihood ratios needed for Bayesian inference \cite{ramachandran2007bayesian}. Thus, we focus on reformulating the likelihood function to speed up Bayesian IRL. 

An ideal likelihood function would require little computation and minimal interaction with the environment. One promising candidate is to leverage recent work on learning from ranked demonstrations \cite{christiano2017deep,browngoo2019trex,brown2019drex}. Given ranked demonstrations, Brown et al. \cite{browngoo2019trex} proposed the algorithm Trajectory-ranked Reward Extrapolation (T-REX) that performs efficient reward inference by transforming reward function learning into a classification problem using a standard pairwise ranking loss. T-REX removes the need to repeatedly sample from or solve an MDP in the inner loop, allowing IRL to scale to visual imitation learning domains such as Atari. However, T-REX only solves for the maximum likelihood estimate of the reward function. One of our contributions is to show that a similar approach based on a pairwise preference likelihood can allow for efficient sampling from the posterior distribution over reward functions.

We assume that we have a set of $m$ trajectories $D = \{ \tau_1,\ldots,\tau_m \}$ along with a set of pairwise preferences over trajectories $\mathcal{P} =  \{(i,j) : \tau_i \prec \tau_j \}$. Note that we do not require a total-ordering over trajectories. These preferences may come from a human demonstrator or could be automatically generated by watching a learner improve at a task \cite{browngoo2019trex} or via noise injection \cite{brown2019drex}. Some trajectory pairs may not have preference information and some trajectories maybe equally preferred, i.e. $(i,j)$ and $(j,i)$ may both be in set $\mathcal{P}$. The benefit of pairwise preferences over trajectories is that we can now leverage a pair-wise ranking loss to compute the likelihood of a $\mathcal{P}$ given a parameterized reward function hypothesis $R_\theta$. We use the standard Bradley-Terry model \cite{bradley1952rank}, alternatively called the  Luce's choice axiom \cite{luce2012individual}, to obtain the following pairwise ranking likelihood function: 
\begin{equation}\label{eqn:pairwiserank}
P(\mathcal{P}, D  \mid R_\theta) = \prod_{(i,j) \in \mathcal{P}} \frac{e^{\beta R_\theta(\tau_j)}}{e^{\beta R_\theta(\tau_i)} + e^{\beta R_\theta(\tau_j)}},
\end{equation}
where $R_\theta(\tau) = \sum_{s \in \tau} R_\theta(s)$ is the predicted return of trajectory $\tau$ under the reward function $R_\theta$, and $\beta$ is the inverse temperature parameter that models the confidence in the preference labels.

Note that using the likelihood function defined in Equation (\ref{eqn:pairwiserank}) does not require solving an MDP. In fact, it does not require any rollouts or access to the MDP. All that is required is that we first calculate the return of each trajectory under $R_\theta$. We then compare the relative predicted returns to the preference labels to determine the likelihood of the demonstrations under the reward hypothesis $R_\theta$. Given this preference-based likelihood function we can perform preference-based Bayesian reward learning using standard MCMC.

\subsection{Optimizations via Successor Features} \label{sec:optimizations}
B-REX uses a deep network to represent the reward function $R_\theta$. However, MCMC proposal generation and mixing time can be slow if there are many demonstrations and if the network is especially large. To make B-REX more efficient and practical, we propose to limit the proposal to only change the last layer of weights in $R_\theta$ when generating MCMC proposals---we will discuss pretraining $R_\theta$ in a later section. We freeze all but the last layer of weights and use the activations of the penultimate layer as our reward features $\phi(s)$. This allows us to represent the reward at a state as a linear combination of features $R_\theta(s) = w^T \phi(s)$. There are two advantages to this formulation: (1) the proposal dimension for MCMC is significantly reduced, allowing for faster convergence; (2) we can efficiently compute the expected value of a policy via a single dot product, and (3) the computation required to calculated the proposal likelihood is significantly reduced.

Given $R(s) = w^T \phi(s)$, the value function of a policy can be written as \begin{equation}
V^\pi_R = \mathbb{E}_{\pi}[\sum_{t=0}^T R(s_t)] = \mathbb{E}_{\pi}[\sum_{t=0}^T w^T \phi(s_t)] = w^T \mathbb{E}_{\pi}[\sum_{t=0}^T \phi(s_t)] = w^T \Phi_\pi,
\end{equation}
where we assume a finite horizon MDP with horizon $T$ and where $\Phi_\pi$ are the successor features \cite{barreto2017successor} of $\pi$. Given any evaluation policy $\pi_{\rm eval}$, we can compute the successor feature once to obtain, $\Phi_{\rm eval}$. We can then compute the expected value of $\pi_{\rm eval}$ as $w^T \Phi_{\pi_{\rm eval}}$ for any reward function weights, $w$.

Using a linear combination of features also allows us to efficiently compute the pairwise ranking losses in the likelihood function. Consider the likelihood function in Equation~(\ref{eqn:pairwiserank}). A naive computation of the likelihood would require $O(T \cdot |\mathcal{P}|)$ forward passes through the deep neural network $R_\theta$ per proposal evaluation, where $|\mathcal{P}|$ is the number of pairwise preferences over demonstration trajectories and $T$ is the length of the trajectories. Given that we would like to potentially generate thousands of samples from the posterior distribution over reward functions, this can significantly slow down MCMC. However, we can reduce this computational cost by noting that 
\begin{equation}
R_\theta (\tau) = \sum_{s \in \tau} w^T\phi(s) = w^T\sum_{s \in \tau} \phi(s) = w^T \Phi_{\tau}.
\end{equation}
Thus, we can precompute and cache $\Phi_{\tau_i} = \sum_{s \in \tau_i} \phi(s)$ for $i = 1,\ldots,m$. The likelihood can then be quickly evaluated as
\begin{equation}\label{eqn:lincombo_Boltzman}
P(\mathcal{P}, D  \mid R_\theta) = \prod_{(i,j) \in \mathcal{P}} \frac{e^{\beta w^T \Phi_{\tau_j}}}{e^{\beta w^T \Phi_{\tau_j}} + e^{\beta w^T \Phi_{\tau_i}}}.
\end{equation}
This results in only $O(|\mathcal{P}|)$ dot products per proposal, resulting in a significant computational savings when generating long MCMC chains over deep neural networks.

When we refer to B-REX in the remainder of this paper we will refer to the optimized version described in this section. See Algorithm~\ref{alg:DeeP-BIRL} in the Appendix for full pseudo-code.
 We found that generating 100,000 reward hypothesis for Atari imitation learning tasks takes approximately 5 minutes on a Dell Inspiron 5577 personal laptop with an Intel i7-7700 processor and an NVIDIA GTX 1050 GPU. In comparison, using standard Bayesian IRL to generate \textit{one sample} from the posterior takes 10+ hours of training for a parallelized PPO reinforcement learning agent  \cite{schulman2017proximal,baselines}.

\subsection{Pretraining the Reward Function Network} \label{sec:pretraining}
Precompute the successor features $\Phi_\tau$ assumes that we already know a good $\phi(s)$.
But how do we train $\phi(s)$ from raw visual features? 
One way is to pretrain $R_\theta$ using T-REX \cite{browngoo2019trex} to find the weight parameters that result in a maximum likelihood estimate given the rankings. Then we can freeze all but the last layer of weights and perform MCMC. Another option is to train the network using an auxiliary loss. Possible candidate auxiliary losses are (1) an inverse dynamics model that uses embeddings $\phi(s_t)$ and $\phi(s_{t+1})$ to predict the corresponding action $a_t$ \cite{torabi2018behavioral,hanna2017grounded}, (2) a variational pixel-to-pixel autoencoder where $\phi(s)$ is the learned latent encoding \cite{makhzani2017pixelgan,doersch2016tutorial}, (3) a cross-entropy loss to learn an embedding $\phi(s)$ that can be used to classify how many timesteps apart are two randomly chosen frames \cite{imitationyoutube}, and (4) a forward dynamics model that predicts $s_{t+1}$ from $\phi(s_t)$ and $a_t$ \cite{oh2015action,thananjeyan2019extending}.

\subsection{HCOPE-IL via B-REX}
We now discuss how to use B-REX to find solutions to the high-confidence off-policy evaluation for imitation learning (HCOPE-IL) problem (see Section~\ref{sec:hcope-il}) when learning from raw visual demonstrations. 
Given samples from the distribution $P(w | D, \mathcal{P})$, where $R(s) = w^T \phi(s)$, we can compute the posterior distribution over any performance statistic $g(\pi_{\rm eval}, R^*)$ as follows. For each sampled weight vector $w$ produced by B-REX, we compute $g(\pi_{\rm eval}, w)$. This results in a sample from the posterior distribution $P(g(\pi_{\rm eval}, R) | \mathcal{P}, D)$, the posterior distribution over performance statistic $g$ conditioned on $D$ and $\mathcal{P}$.  We then compute a $(1-\delta)$ confidence lower bound, $\hat{g}(\pi_{\rm eval}, D)$, by finding the $\delta$-quantile of $g(\pi_{\rm eval}, w)$ for $w \sim P(w|\mathcal{P}, D)$. In our experiments we focus on bounding the expected value of the evaluation policy, i.e., $g(\pi_{\rm eval}, R^*) = V^{\pi_{\rm eval}}_{R^*} = {w^*}^T \Phi_{\pi_{\rm eval}}$. To compute a $1 - \delta$ confidence bound on $V^{\pi_{\rm eval}}_{R^*}$, we take full advantage of the successor feature representation to efficiently calculate the posterior distribution over policy returns given preferences and demonstrations via a simple matrix vector product,  $W \Phi_{\pi_{\rm eval}}$,
where each row of $W$ is a sample, $w$, from the MCMC chain and $\pi_{\rm eval}$ is the evaluation policy. We then sort the elements of this vector and select the $\delta$-quantile. This gives us a $1-\delta$ confidence lower bound on $V^{\pi_{\rm eval}}_{R^*}$ and corresponds to calculating the $\delta$-Value at Risk (VaR) over $V^{\pi_{\rm eval}}_{R}\sim P(R | \mathcal{P}, D)$ \cite{brown2018efficient,jorion1997value,tamar2015optimizing}. 

\section{Experimental Results}

\subsection{Imitation Learning via B-REX}
We first tested the efficacy of B-REX to see if it can be used to find a reward function that leads to good policies via reinforcement learning. We enforce constraints on the weight vectors by normalizing the output of the weight vector proposal such that $\|w\|_1 = 1$ and use a Gaussian proposal function centered on $w$ with standard deviation $\sigma$. Thus, given the current sample $w_t$, the proposal is defined as $w_{t+1} = \texttt{normalize}(\mathcal{N}(w_t, \sigma))$,
where \texttt{normalize} projects the sample back to the surface of the L1-unit ball. We used models pretrained from pairwise preferences using T-REX to obtain $\phi(s)$ \cite{browngoo2019trex}.\footnote{Pretrained networks are available at \url{https://github.com/hiwonjoon/ICML2019-TREX/tree/master/atari/learned_models/icml_learned_rewards}} This results in a 65 dimensional features vector $\phi(s)$. 
We ran MCMC for 100,000 steps with $\sigma = 0.005$ and with a uniform prior. Due to our proposed optimizations this only required a few minutes of computation time. We then took the MAP and mean reward estimates and optimized a policy using Proximal Policy Optimization \cite{schulman2017proximal}.

\begin{table}
  \caption{Ground-truth average returns for several Atari games when optimizing the mean and MAP rewards found using B-REX. We also compare against reported results for T-REX \cite{browngoo2019trex}. Each algorithm is given the same 12 demonstrations with ground-truth pairwise preferences. The average performance for each IRL algorithm is the average over 30 rollouts.}
  \label{tab:deepbirl_rl}
  \centering
    \vspace{0.1cm}
\begin{tabular}{ccccccc}
\toprule
 & \multicolumn{2}{c}{Ranked Demonstrations} & \multicolumn{1}{c}{B-REX Mean} & \multicolumn{1}{c}{B-REX MAP} & T-REX \\
 \midrule
Game &  Best & Average & Average & Average & Average\\ 
\midrule 
Beam Rider &	1332 &	686.0 &	 878.7 &
 1842.6  & \textbf{3,335.7}\\
Breakout &	32 &	14.5 &  392.5 & \textbf{419.7}	 & 221.3\\
Enduro &	84 &	39.8 & 450.1
 & 569.7 & \textbf{586.8}\\
Seaquest &	600 &	373.3 &  \textbf{967.3}  & 570.7 & 747.3\\ 
Space Invaders &	600 &	332.9 & 1437.5 & \textbf{1440.2} & 1,032.5\\ 
\bottomrule
\end{tabular}
\end{table}

We tested our approach on five Atari games from the Arcade Learning Environment \cite{bellemare2013arcade}. Because we are concerned with imitation learning, we mask game scores and life information and the imitation learning agent does not receive the ground-truth reward signal. All that is available are pairwise preferences on state trajectories.
Table~\ref{tab:deepbirl_rl} shows results of performing RL on the mean and MAP rewards found using B-REX. We ran all algorithms using the same demonstrations and preference labels. We used the same 12 suboptimal demonstrations used by Brown et al. and give each algorithm all pairwise preference labels based on the ground-truth returns. T-REX uses a sigmoid to normalize rewards before passing them to the RL algorithm; however, we obtained better performance for B-REX by feeding the unnormalized predicted reward $R_\theta(s)$ into PPO for policy optimization.

Table~\ref{tab:deepbirl_rl} shows that, similar to T-REX, B-REX is able to utilize preferences to outperform the demonstrator. B-REX is competitive with T-REX, achieving better average scores on 3 out of 5 games. Additionally, we found that using the MAP reward from the posterior was superior to optimizing for the mean reward on 4 out of 5 games. We also found that B-REX can successfully use pairwise preferences over suboptimal demonstrations to learn a better-than-demonstrator policy. When optimizing for the mean reward, B-REX is able to obtain an average performance that surpasses the performance of the best demonstration in every game except for Beam Rider. When optimizing for the MAP reward, B-REX is obtains an average performance that surpasses the best demonstration on all games, except for Seaquest.

\subsection{High-Confidence Lower Bounds on Policy Performance}
\begin{figure}
 \begin{subfigure}{.5\textwidth}
 \centering
\includegraphics[width=.8\linewidth]{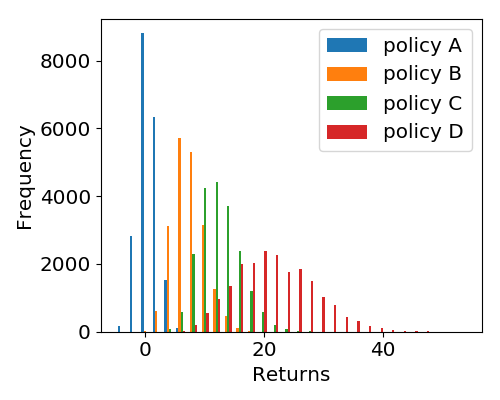}
 \caption{Posterior}
  \label{}
\end{subfigure}
 \begin{subfigure}{.5\textwidth}
 \centering
\includegraphics[width=.8\linewidth]{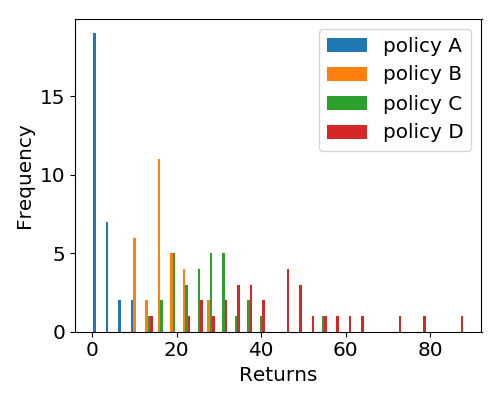}
 \caption{Ground Truth}
  \label{}
\end{subfigure}
\caption{Breakout return distributions over the posterior $P(R | D, \mathcal{P})$ compared with ground truth game scores. Policies A-D correspond to checkpoints of an RL policy partially trained on the ground-truth reward function and correspond to 25 (A), 325 (B), 800 (C), and 1450 (D) training updates to PPO. The learned posterior distributions roughly match the general shapes of the true distribution.}
\label{fig:breakout_distplots}
\end{figure}

Next we ran an experiment to validate whether the posterior distribution generated by B-REX can be used for accurately bounding the expected return of the evaluation policy under the unknown reward function $R^*$. 
We estimated $\Phi_{\pi_{\rm eval}}$ using 30 Monte Carlo rollouts. We first evaluated four different evaluation policies, $A \prec B \prec C \prec D$, created by partially training a PPO agent on the ground-truth reward function. We ran B-REX to generate 100,000 samples from $P(R | \mathcal{P}, D)$. Figure~\ref{fig:breakout_distplots} shows predicted and ground truth distributions for the four different evaluation policies: A--D. We found that the predicted distributions (100,000 MCMC samples) have roughly similar shape to the ground truth distribution (30 rollouts). Note we do not know the scale of the true reward $R^*$. Thus, the results from B-REX are most useful when comparing the relative performance of several different evaluation policies \cite{brown2018efficient}. We see that the modes predicted by B-REX match the ordering of the modes of policies A--D under the true reward function.

\begin{table}
  \caption{Policy evaluation statistics for Breakout over the  return distribution from the learned posterior $P(R | D, \mathcal{P})$ compared with the ground truth returns using game scores. Policies A-D correspond to checkpoints of an RL policy partially trained on the ground-truth reward function and correspond to 25, 325, 800, and 1450 training updates to PPO. The mean and MAP policies are the results of PPO using the mean and MAP rewards, respectively. No-Op is a policy that never takes the action to release the ball, resulting in a lower 0.05-quantile return (0.05-VaR).}
  \label{tab:breakoutPolicyEval}
  \centering
  \vspace{0.1cm}
  \begin{tabular}{crrrrr}
    \toprule
 Policy & Mean Chain & 0.05-VaR Chain & Traj. Length & GT Avg. Return & GT Min. Return \\
 \midrule
policy A & 0.8 & -1.7 & 213.8 & 2.2 & 0  \\
policy B & 7.4 & 3.6 & 630.1 & 16.6 & 9  \\
policy C & 12.4 & 7.5 & 834.5 & 26.7 & 12  \\
policy D & 21.5 & 11.6 & 1070.6 & 43.8 & 14  \\
mean & 88.1 & 25.2 & 3250.4 & 392.5 & 225  \\
MAP & 2030.4 & 75.7 & 29761.4 & 419.7 & 261  \\
No-Op & 6256.2 & -134.9 & 99994.0 & 0.0 & 0  \\
    \bottomrule
  \end{tabular}
\end{table}

Table~\ref{tab:breakoutPolicyEval} shows the numerical results for evaluating policies under $P(R \mid D, \mathcal{P})$. We show results for partially trained policies A-D as well as well as policies trained on the MAP reward, the mean reward, and a No-Op policy. We found that the ground-truth returns for the checkpoints were highly correlated with the mean reward found under the chain and the 0.05-VaR (5th percentile policy return) under the chain. However, we also noticed that the trajectory length was also highly correlated with the ground-truth reward. If the reward function learned via IRL gives a small positive reward at every timestep, then long polices that do the wrong thing may look good under the posterior. To test this we evaluated a No-Op policy that seeks to hack the learned reward function by never releasing the ball in Breakout. We ran the No-Op policy until the Atari emulator timed out after 99,994 no-ops.

The bottom row of Table~\ref{tab:breakoutPolicyEval} shows that while the No-Op policy has a high expected return over the chain, looking at the 0.05-VaR shows that the No-Op policy has high risk under the distribution, much lower than even policy A which on average only scores 2.0 points. This finding validates the results by Brown and Niekum \cite{brown2018efficient} that demonstrated the value of using a probabilistic worst-case bound for evaluating the performance of policies when the true reward function is unknown. Our results demonstrate that reasoning about probabilistic worst-case performance may be one potential way to detect policies that have overfit to certain features in the demonstrations that are correlated with the intent of the demonstrations, but do not lead to desired behavior, the so-called reward hacking problem \cite{amodei2016concrete}. See the Appendix for results for all games. We found that for some of the games, the learned posterior is not as useful for accurately ranking policies. We hypothesize that this may be because the pretrained features $\phi(s)$ are overfit to the rankings. In the future we hope to improve these results by using additional auxiliary losses when pretraining the reward features (see Section~\ref{sec:pretraining}).

\section{Summary}
Bayesian reasoning is a powerful tool when dealing with uncertainty and risk; however, existing Bayesian inverse reinforcement learning algorithms require solving an MDP in the inner loop, rendering them intractable for complex problems where solving an MDP may take several hours or even days. In this paper we propose a novel deep learning algorithm, Bayesian Reward Extrapolation (B-REX), that leverages preference labels over demonstrations to make Bayesian IRL tractable for high-dimensional visual imitation learning tasks. B-REX can sample tens of thousands of reward functions from the posterior in a matter of minutes using a consumer laptop. We tested our approach on five Atari imitation learning tasks and demonstrated B-REX is competitive with state-of-the-art imitation learning methods. Using the posterior samples produced by B-REX, we demonstrated for the first time that it is computationally feasible to compute high-confidence performance bounds for arbitrary evaluation policies given demonstrations of visual imitation learning tasks. Our proposed bounds can allow accurate comparison of different evaluation policies and provide a potential way to detect reward hacking.
In the future we are interested in using high-confidence bounds on policy performance to implement safe and robust policy improvement in the imitation learning setting. Given a starting policy $\pi$ we want to optimize a policy such that it maximizes some safety threshold. 
One possible way to improve the policy would be to use an evolutionary strategy where the fitness is simply the lower bound on the performance metric calculated over the posterior distribution of reward functions. 
We also plan to experiment with different architectures and different pretraining schemes for learning reward features automatically from raw visual features. 

\bibliographystyle{plain}
\bibliography{deepbirl}

\appendix

\section{Preference-based Bayesian IRL}
Pseudo-code for the optimized version of Bayesian Reward Extrapolation (B-REX) is shown in Algorithm~\ref{alg:DeeP-BIRL}.

\begin{algorithm}[t]
\caption{B-REX: Bayesian Reward Extrapolation}
\label{alg:DeeP-BIRL}
\begin{algorithmic}[1]
\State \textbf{input:} demonstrations $D$, pairwise preferences $\mathcal{P}$, MCMC proposal width $\sigma$, number of proposals to generate $N$, deep network architecture $R_\theta$, and prior $P(w)$.
\State Pretrain $R_\theta$ using auxiliary tasks (see Section~\ref{sec:pretraining}).
\State Freeze all but last layer, $w$, of $R_\theta$. Let $\phi(s)$ be the activations of the penultimate layer of $R_\theta$.
\State Precompute and cache $\Phi_{\tau} = \sum_{s \in \tau} \phi(s)$ for all $\tau \in D$.
\State Initialize $w$ randomly.
\State Chain[0] $\gets  w$
\State Compute $P(\mathcal{P}, D | w) P(w)$ using Equation~(\ref{eqn:lincombo_Boltzman})
\For{$i \gets 1$ to $N$}
	\State $\tilde{w} \gets \mathcal{N}(w,\sigma)$	\Comment 	Normal proposal distribution
	\State Compute $P(\mathcal{P}, D | \tilde{w}) P(\tilde{w})$ using Equation~(\ref{eqn:lincombo_Boltzman})
		\State	$u \gets \texttt{Uniform} (0,1)$
	\If {$\displaystyle u < \frac{P(\mathcal{P}, D | \tilde{w}) P(\tilde{w})}{P(\mathcal{P}, D | w) P(w)}$} 
    	\State Chain[i] $\gets \tilde{w}$
    	\vspace{.5mm}
    	\State $ w \gets \tilde{w}$
	\Else
		\State Chain[i] $\gets w$
	\EndIf
\EndFor
\State \textbf{return} Chain
\end{algorithmic}
\end{algorithm}

\section{Mixing plots}
\begin{figure}
 \begin{subfigure}{.33\textwidth}
 \centering
\includegraphics[width=.9\linewidth]{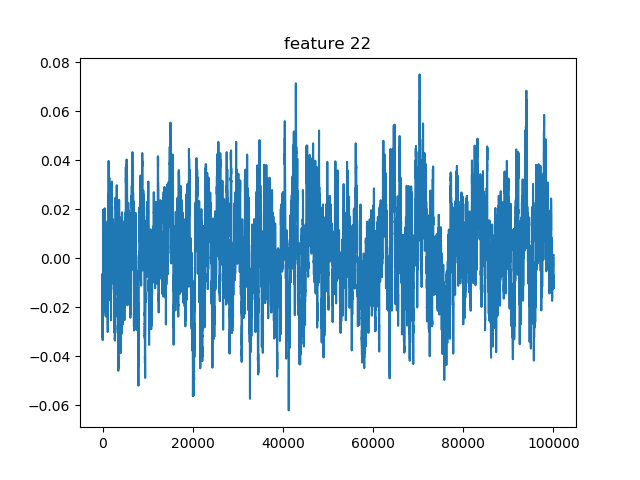}
 \caption{}
  \label{subfig:feature22}
\end{subfigure}
 \begin{subfigure}{.33\textwidth}
 \centering
\includegraphics[width=.9\linewidth]{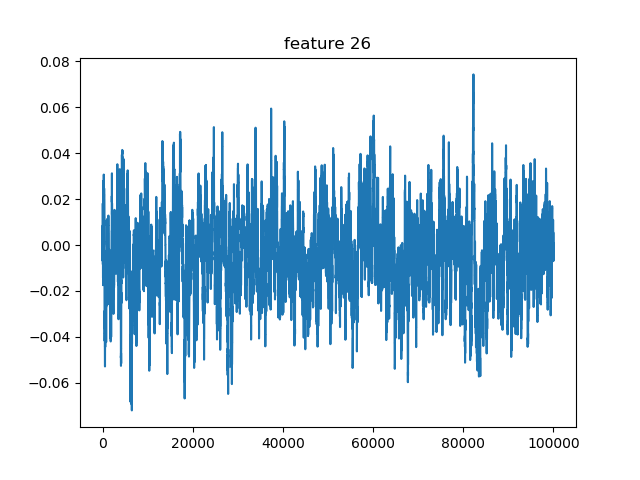}
 \caption{}
  \label{subfig:feature26}
\end{subfigure}
 \begin{subfigure}{.33\textwidth}
 \centering
\includegraphics[width=.9\linewidth]{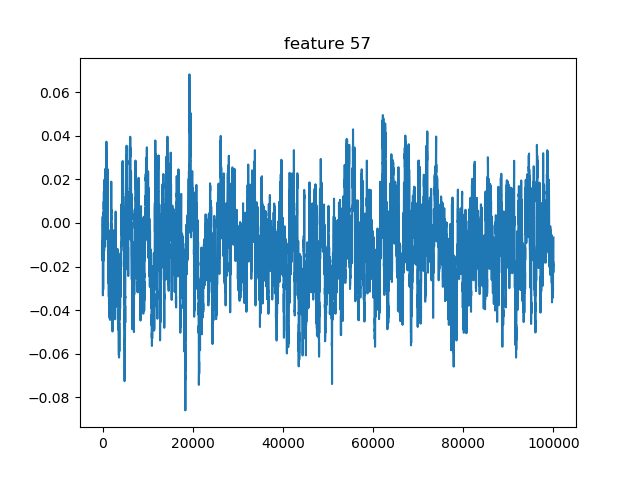}
 \caption{}
  \label{subfig:feature57}
\end{subfigure}
\caption{MCMC mixing plots for three randomly chosen weight features. Total dimensionality of weight vector is 65. MCMC was performed for 100,000 steps with a proposal width of $\sigma = 0.005$. Weights are normalized so that $\|w\|_1 = 1$.}
\label{fig:mcmc_mixing}
\end{figure}

The mixing plots for three randomly chosen features for Breakout are shown in Figure~\ref{fig:mcmc_mixing} and appear to be rapidly mixing.

\section{Plots of return distributions for MCMC chain and ground-truth rewards}
\begin{figure}
 \begin{subfigure}{.5\textwidth}
 \centering
\includegraphics[width=.99\linewidth]{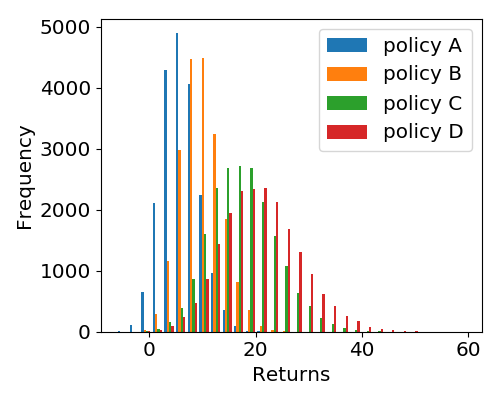}
 \caption{Posterior}
  \label{}
\end{subfigure}
 \begin{subfigure}{.5\textwidth}
 \centering
\includegraphics[width=.99\linewidth]{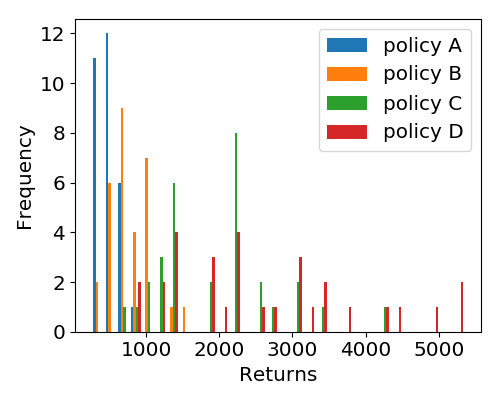}
 \caption{Ground Truth}
  \label{}
\end{subfigure}
\caption{Beam Rider return distributions over the posterior $P(R | D, \mathcal{P})$ compared with the ground truth returns using game scores. Policies A-D correspond to checkpoints of an RL policy partially trained on the ground-truth reward function and correspond to 25, 325, 800, and 1450 training updates to PPO. }
\label{fig:beamrider_distplots}
\end{figure}

\begin{figure}
 \begin{subfigure}{.5\textwidth}
 \centering
\includegraphics[width=.99\linewidth]{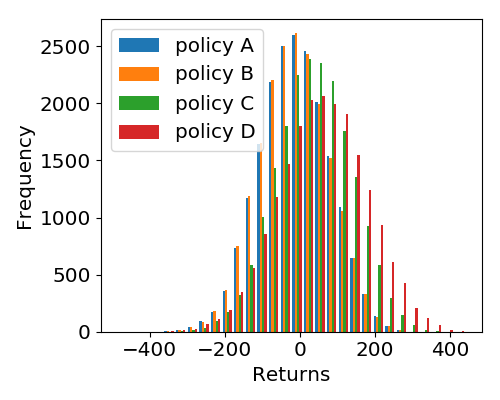}
 \caption{Posterior}
  \label{}
\end{subfigure}
 \begin{subfigure}{.5\textwidth}
 \centering
\includegraphics[width=.99\linewidth]{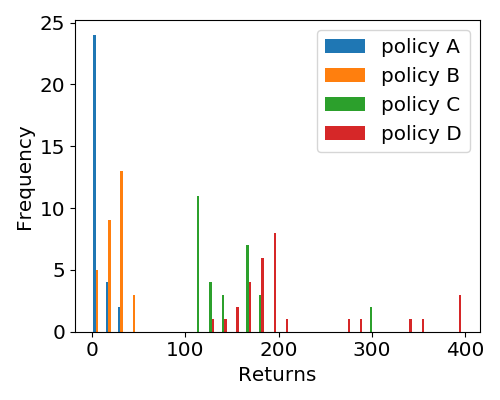}
 \caption{Ground Truth}
  \label{}
\end{subfigure}
\caption{Enduro return distributions over the posterior $P(R | D, \mathcal{P})$ compared with the ground truth returns using game scores. Policies A-D correspond to checkpoints of an RL policy partially trained on the ground-truth reward function and correspond to 3125, 3425, 3900, 4875 training updates to PPO. }
\label{fig:enduro_distplots}
\end{figure}

\begin{figure}
 \begin{subfigure}{.5\textwidth}
 \centering
\includegraphics[width=.99\linewidth]{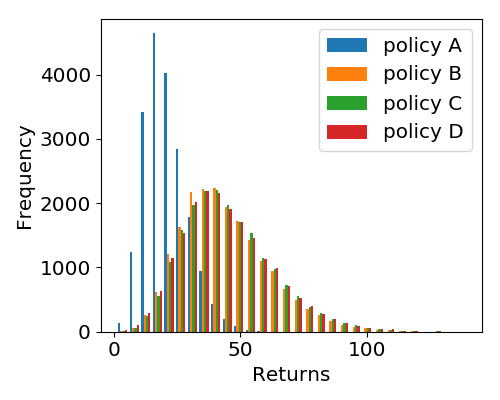}
 \caption{Posterior}
  \label{}
\end{subfigure}
 \begin{subfigure}{.5\textwidth}
 \centering
\includegraphics[width=.99\linewidth]{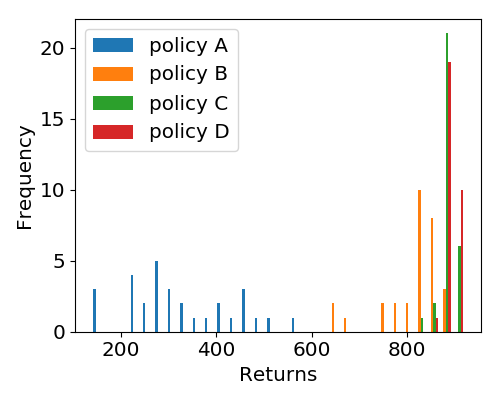}
 \caption{Ground Truth}
  \label{}
\end{subfigure}
\caption{Seaquest return distributions over the posterior $P(R | D, \mathcal{P})$ compared with the ground truth returns using game scores. Policies A-D correspond to checkpoints of an RL policy partially trained on the ground-truth reward function and correspond to 25, 325, 800, and 1450 training updates to PPO. }
\label{fig:seaquest_distplots}
\end{figure}

\begin{figure}
 \begin{subfigure}{.5\textwidth}
 \centering
\includegraphics[width=.99\linewidth]{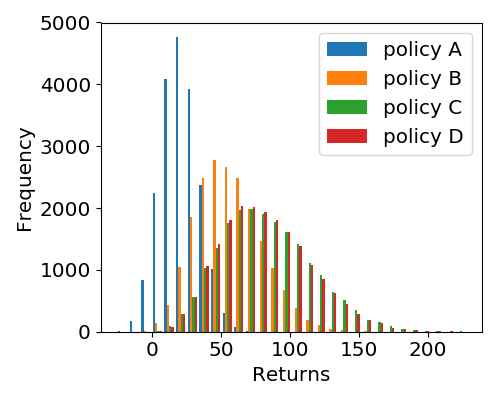}
 \caption{Posterior}
  \label{}
\end{subfigure}
 \begin{subfigure}{.5\textwidth}
 \centering
\includegraphics[width=.99\linewidth]{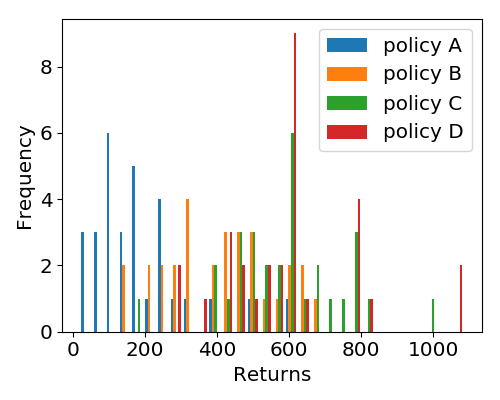}
 \caption{Ground Truth}
  \label{}
\end{subfigure}
\caption{Space Invaders return distributions over the posterior $P(R | D, \mathcal{P})$ compared with the ground truth returns using game scores. Policies A-D correspond to checkpoints of an RL policy partially trained on the ground-truth reward function and correspond to 25, 325, 800, and 1450 training updates to PPO. }
\label{fig:spaceinvaders_distplots}
\end{figure}

Figures~\ref{fig:beamrider_distplots}--\ref{fig:spaceinvaders_distplots} show the predicted and ground-truth distributions for different evaluation policies for Beam Rider, Enduro, Seaquest, and Space Invaders.

\section{High-Confidence Lower Bounds (Full results)}
We report the full results for computing high-confidence lower bounds on policy performance for Beam Rider, Enduro, Seaquest, and Space Invaders in tables~\ref{tab:beamriderPolicyEval}--\ref{tab:spaceinvadersPolicyEval}.

\begin{table}
  \caption{Policy Evaluation for Beam Rider.}
  \label{tab:beamriderPolicyEval}
  \centering
  \begin{tabular}{crrrrr}
    \toprule
 Policy & Mean Chain & 0.05-VaR Chain & Traj. Length & GT Avg. Return & GT Min. Return \\
 \midrule
policy A & 6.2 & 0.8 & 1398.8 & 471.3 & 264  \\
policy B & 9.9 & 4.2 & 1522.9 & 772.3 & 396  \\
policy C & 17.6 & 8.2 & 2527.5 & 1933.3 & 660  \\
policy D & 20.7 & 9.6 & 2963.2 & 2618.5 & 852  \\
mean & 97.4 & 61.2 & 8228.4 & 878.7 & 44  \\
MAP & 76.7 & 46.5 & 7271.0 & 1842.6 & 264  \\
    \bottomrule
  \end{tabular}
\end{table}

\begin{table}
  \caption{Policy Evaluation for Enduro.}
  \label{tab:enduroPolicyEval}
  \centering
  \begin{tabular}{crrrrr}
    \toprule
 Policy & Mean Chain & 0.05-VaR Chain & Traj. Length & GT Avg. Return & GT Min. Return \\
 \midrule
policy A & -12.0 & -163.0 & 3322.1 & 7.7 & 0  \\
policy B & -12.9 & -163.9 & 3322.1 & 25.8 & 1  \\
policy C & 35.3 & -127.9 & 3544.0 & 149.2 & 106  \\
policy D & 56.7 & -132.8 & 4098.7 & 215.6 & 129  \\
mean & 164.1 & -156.0 & 6761.1 & 450.1 & 389  \\
MAP & 208.6 & -176.5 & 8092.3 & 569.7 & 441  \\
    \bottomrule
  \end{tabular}
\end{table}

\begin{table}
  \caption{Policy Evaluation for Seaquest.}
  \label{tab:seaquestPolicyEval}
  \centering
  \begin{tabular}{crrrrr}
    \toprule
 Policy & Mean Chain & 0.05-VaR Chain & Traj. Length & GT Avg. Return & GT Min. Return \\
 \midrule
policy A & 21.3 & 9.6 & 1055.9 & 327.3 & 140  \\
policy B & 43.9 & 19.8 & 2195.7 & 813.3 & 640  \\
policy C & 45.0 & 20.3 & 2261.8 & 880.7 & 820  \\
policy D & 44.6 & 19.3 & 2265.5 & 886.7 & 860  \\
mean & 63.1 & 27.5 & 2560.0 & 967.3 & 740  \\
MAP & 56.8 & 24.1 & 2249.1 & 570.7 & 520  \\
    \bottomrule
  \end{tabular}
\end{table}

\begin{table}
  \caption{Policy Evaluation for Space Invaders.}
  \label{tab:spaceinvadersPolicyEval}
  \centering
  \begin{tabular}{crrrrr}
    \toprule
 Policy & Mean Chain & 0.05-VaR Chain & Traj. Length & GT Avg. Return & GT Min. Return \\
 \midrule
policy A & 21.5 & -0.4 & 527.3 & 186.0 & 20  \\
policy B & 55.8 & 20.4 & 715.5 & 409.2 & 135  \\
policy C & 82.3 & 34.0 & 876.7 & 594.0 & 165  \\
policy D & 81.2 & 34.0 & 824.7 & 602.3 & 270  \\
mean & 257.2 & 113.6 & 2228.7 & 1437.5 & 515  \\
MAP & 246.2 & 109.0 & 2100.4 & 1440.2 & 600 \\
    \bottomrule
  \end{tabular}
\end{table}

\end{document}